\documentclass[letterpaper]{article}

\usepackage{natbib,alifeconf}  
\usepackage{svg}
\usepackage{amsmath}
\usepackage{url,hyperref,cleveref}
\usepackage{booktabs}
\usepackage{caption}
\usepackage{tikz}
\usetikzlibrary{arrows.meta, positioning}

\newcommand\blfootnote[1]{%
  \begingroup
  \renewcommand\thefootnote{}\footnote{#1}%
  \addtocounter{footnote}{-1}%
  \endgroup
}

\title{Life as Plasmas: Autonomy and Interactivism \emph{in-materio}}

\author{
    Nicolás Hinrichs$^{1,6}$,
    Mahault Albarracin$^{2}$,   
    Felipe Engelberger$^{3,6}$, \\
    \Large Leonardo Christov-Moore $^{4}$, \and
    Daniel Polani$^{5}$ \\
    \mbox{}\\
    $^1$Max Planck Institute for Human Cognitive and Brain Sciences, Germany \\
    $^2$Verses AI Research Lab / Université du Québec à Montréal, Canada \\
    $^3$Institute for Drug Discovery, Universität Leipzig, Germany \\
    $^{4}$Institute for Advanced Consciousness Studies, USA \\
    $^{5}$University of Hertfordshire, UK \\
    $^{6}$ Eigenify, USA\\
}

\begin{document}
 
\maketitle
 
\begin{abstract}
When is a material system a candidate for life at all? We argue that this question is prior to behavior, functional architecture, or computational capacity, and that at root it is one of physical admissibility. We develop a framework in which minimal autonomy, taken in the interactivist sense of normativity grounded in self-maintaining far-from-equilibrium organization, corresponds to a distinct non-equilibrium phase of matter, and we take complex plasmas, a physical and non-biological system, as its \emph{in-materio} exemplar. We formalize a diagnostic phase-space whose criteria (sustained free-energy throughput, organizational closure, active information maintenance, and regulated noise sensitivity) constitute necessary conditions for life-attribution. We instantiate the diagnostics across contrasting systems and fix the boundaries of the phase space via Bénard convection as a driven baseline lacking closure, and a digital self-replicating soup that carries measured informational heredity while its physical closure remains a structural zero. We demonstrate that  plasmas satisfy every admissibility condition for minimal physical autonomy while carrying none of the informational heredity that open-ended evolution requires, sharpening the distinction between physical admissibility and biological sufficiency, and bounding questions of machine sentience.
\end{abstract}

Data/Code available at: \url{https://doi.org/10.5281/zenodo.20732279}
\blfootnote{\textcopyright  2026 Hinrichs, Albarracin, Engelberger, Christov-Moore, and Polani. Published under a Creative Commons Attribution 4.0 International (CC BY 4.0) license. \href{mailto:hinrichsn@cbs.mpg.de}{hinrichsn@cbs.mpg.de}}

\section{Introduction}
 
Can life and cognition be abstracted into software, or does autonomy require a thermodynamic substrate? How far can we push the substrate? Must life be organic, instantiated, physical, or can it be inorganic, simulated, and computational?
 
Strict computational substrate independence  has been challenged by the normative view that living systems, embodied and thermodynamicaly precarious, have a capacity for ``sense-making'' rooted in its physical struggle against entropic disorder \citep{Thompson2007}; purely syntactic computation might lack the physical dynamics necessary for genuine agency. 
 
Contemporary debates about machine consciousness encounter a structurally identical problem. Accounts that evaluate candidate systems through behavior, architectural motifs, or abstract information-theoretic metrics presuppose a physical substrate capable of sustaining the organized dynamics those metrics are meant to capture. Integrated Information Theory \citep{Tononi2016} evaluates intrinsic cause-effect power; a criterion that, as we argue below, supports physical grounding and does not license substrate-independent attribution. Global Workspace Theory \citep{Baars1988} similarly presupposes an organizational architecture whose physical realization is left unexamined.

 \begin{figure}[h]
\centering
\resizebox{\columnwidth}{!}{%
\begin{tikzpicture}[
    node distance=1.8cm and 1.8cm,
    every node/.style={
        align=center,
        font=\footnotesize,
        text width=3.4cm
    },
    arrow/.style={->, dashed}
]
\node (substrates) {
\textbf{Driven Physical Substrates}\\
plasmas, embodied hardware, active matter
};
\node (organization) [right=of substrates] {
\textbf{Non-Equilibrium}\\
\textbf{Organized Systems}\\
sustained throughput,\\
multiscale dynamics
};
\node (diagnostics) [below=of substrates] {
\textbf{Diagnostic Phase Space}\\
$\{Q,\;\sigma,\;\Phi_{\mathrm{closure}},\;R_{\mathrm{info}},\;\chi_{\mathrm{noise}}\}$
};
\node (attribution) [right=of diagnostics] {
\textbf{Admissible Attribution Domain}\\
life- or consciousness-relevant systems
};
\draw[arrow] (substrates) -- (organization);
\draw[arrow] (diagnostics) -- (attribution);
\draw[arrow] (substrates) -- (diagnostics);
\draw[arrow] (organization) -- (attribution);
\end{tikzpicture}
}
\vspace{-0.4em}
\caption{\textbf{Commutative grounding and attribution.}
Life- or autonomy-attribution is admissible only when both paths from \emph{Driven Physical Substrates} to \emph{Admissible Attribution Domain} agree.}
\label{fig:commutative-attribution}
\end{figure}
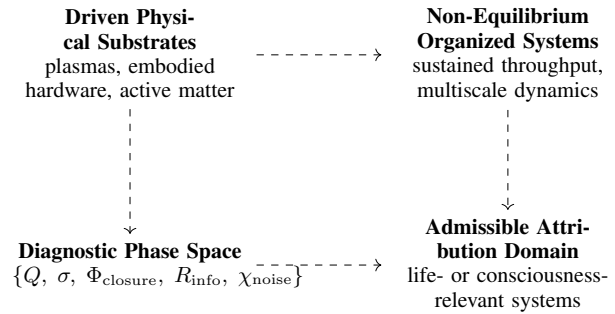

We ask: \emph{what kinds of material systems are even capable of sustaining the organizational conditions that make life and consciousness plausible?} On this view, the prior question for mind and life is whether a system occupies the same far-from-equilibrium organizational state as living systems, and only then whether it exhibits intelligence or representation. This sustained far-from-equilibrium stance, together with recursive self-maintenance and information use, has motivated the view that life constitutes a distinct \emph{state of matter} \citep{Schrodinger:1944, GoldenfeldWoese2011,Bauer1920}, irreducible to any particular molecular arrangement. Figure~\ref{fig:commutative-attribution} formalizes this grounding requirement as a commutative diagram: a substrate must satisfy non-equilibrium organizational criteria (sustained throughput, multiscale dynamics), and its measurable diagnostics $\{Q,\;\sigma,\;\Phi_{\mathrm{closure}},\;R_{\mathrm{info}},\;\chi_{\mathrm{noise}}\}$ must fall within bounded admissible ranges. Debates about (artificial) life, (machine) consciousness, and (collective) intelligence are usually approached through functional behavior or symbolic computation; we here treat one question as prior with regard to their tractability: whether a given substrate can physically sustain the organization those debates presuppose.

To investigate the origins of agency without presupposing biological chemistry, we require non-biological physical models. Recent \emph{ALife in-materio} work searches for life-like behaviour in physical substrates governed by entirely different principles than biology \citep{Penty2025alife}, reducing to neither soft (abstract simulation), hard (robotic embodiment), nor wet (biochemical) \emph{ALife}: the creatures are virtual patterns embedded in the material's collective physical state, yet governed by physics directly. Our work is complementary. We introduce the dusty plasma chamber as a synthetic model in the tradition of experimental philosophy through physical instantiation \citep{DiPaolo2000}, extending it to the thermodynamic question \emph{in-materio} approaches have not formally addressed: which physical conditions are \emph{necessary} for life- or consciousness-relevant organization at all? We deploy macroscopic dusty plasmas as \emph{opaque thought experiments} whose complex non-linear dynamics resist simulative anticipation and must be observed systematically to yield philosophical insights.

In its search-and-characterization aspect this work is closest to ASAL \citep{kumar2025automatingsearchartificiallife}, whose foundation-model evaluator scores ALife simulations through a learned, opaque representation across many substrates, best instantiated as a self-replicating-program soup \citep{alakuijala2024computational} to illustrate a compression-based emergence signal whose every detection traces back to the underlying bytes and lineage.

We do not claim dusty plasmas are alive. We map plasma dynamics against formal operational criteria, \citeauthor{RuizMirazo2004}'s dual-pillar framework distinguishing basic autonomy from open-ended evolution \citep{RuizMirazo2004} and \citeauthor{Moreno2015}'s constraint closure \citep{Moreno2015}, to identify which physical preconditions for life inorganic matter demonstrably satisfies and which remain open. This graduated evaluation is the paper's central contribution. While \citet{Tsytovich2007} proposed that complex plasmas might constitute ``inorganic living matter,'' that hypothesis has been co-opted by fringe literature extrapolating idealized simulations into unsubstantiated claims; we restrict our analysis to formal operational definitions and measurable laboratory parameters.
 
\section{Autonomy, Precariousness, and Non-Equilibrium Organization} \label{sec:framework}

Before analyzing dusty plasmas, we establish the philosophical and thermodynamic criteria against which physical autonomy is judged.

\subsection{Dissipative Structures and Their Limits}

Living systems violate equilibrium assumptions of ergodicity, detailed balance \citep{klein1955principle}, and convergence to stationary macrostates \citep{frigg2016field}: they maintain gradients, cycle matter and energy, and regenerate the conditions of their own persistence \citep{lynn2021broken}. \citet{Prigogine1967} showed that systems driven far from equilibrium produce ``dissipative structures,'' but this is necessary, not sufficient, for life: most stay \emph{organizationally shallow}, persisting only while external conditions are imposed \citep{ruiz2020construction}. \citet{Maturana1980} termed the missing property \textit{autopoiesis}, the self-referential production network separating non-equilibrium systems \emph{with} organizational closure from those \emph{without}. \citet{Lerchner2026} draws a complementary distinction between \emph{vehicle causality} (transitions driven by substrate physics) and \emph{content causality} (driven by intrinsic semantic content): a modelled self-maintaining network advances when voltages cross thresholds, with no threat of dissolution behind the advance, so the existential stake grounding biological normativity \citep{Bickhard2000} is absent from the vehicle's causal chain however faithfully the topology is reproduced.
 
\subsection{Closure and the Work-Constraint Cycle}
\citet{Kauffman2000} introduced the ``work-constraint cycle,'' where energy from non-equilibrium processes is channeled by constraints to perform work that regenerates those same constraints. \citet{Moreno2015} formalized this into \emph{constraint closure}: a causal regime where constraints mutually depend on one another for their existence and regeneration. This mutual dependence is the physical grounding of biological teleology: the parts exist for the sake of the whole as a causal fact grounded in the organization, without metaphysical stipulation imposed from outside.We use \emph{organizational closure} \citep{Mossio2023organization} to denote this regime in which internal processes mutually reinforce one another such that the system actively maintains the conditions of its own persistence. Closure marks the transition from passive dissipation to active self-maintenance, from a flame to a cell.

\citet{Rosen1991} formalized a complementary perspective through (M,R)-systems, arguing that a living organism must be ``closed to efficient causation'': every efficient cause needed must be synthesized internally. This impredicativity renders the system non-fractionable and constitutes a stringent formal criterion.

\subsection{Basic Autonomy and Open-Ended Evolution}

\citet{RuizMirazo2004} provide a dual-pillar operational framework avoiding the epistemological vulnerabilities of descriptive definitions of life \cite{Cleland2012,Machery2012}. \emph{Basic autonomy} requires a system to maintain itself far from equilibrium through an active boundary and an energy-transduction apparatus, internal processes continually regenerating the network that produces them. \emph{Open-ended evolution} requires a historical-collective organization with phenotype-genotype decoupling, instructed by material records whose transmission allows indefinite increase in complexity. The framework thus enables graduated evaluation: a system can satisfy basic autonomy while failing open-ended evolution, occupying a precise position in organizational complexity.

\subsection{Precariousness and Normativity}
Following \citet{Bickhard1993,Bickhard2000} and \citet{DiPaolo2005}, normativity emerges from thermodynamic precariousness. \citet{Bickhard2000} distinguishes \emph{simple} from \emph{recursive} self-maintenance: a candle flame sustains its combustion but cannot alter its dynamics to seek a better environment, whereas recursive self-maintenance shifts between qualitatively different dynamical processes to preserve far-from-equilibrium stability across environmental variation, as a bacterium does when it senses a gradient, tumbles, and redirects its chemotaxis machinery. This is the minimal physical basis for sense-making: the threshold at which the environment becomes \emph{meaningful}, and below which information processing carries no existential stake.
 
\subsection{Information and Noise}
Closure entails \emph{information-in-use}: structured correlations that constrain future dynamics and reduce uncertainty about the system's own states, maintained physically, without symbolic encoding. Without sustained maintenance, information does not matter to continued existence \citep{kolchinsky2018semantic}. \citet{WalkerDavies2013} place the defining transition where ``information gains direct and context-dependent causal efficacy over the matter in which it is instantiated,'' and \citet{PRXLife.2.033009} second this, equating intrinsic motivation with seeking unstable Lyapunov exponents in controllable directions only.

Under sustained drive and closure, stochastic fluctuations can actively trigger transitions between viable states, and do more than degrade structure \citep{Tsimring2014}. This sensitivity must be regulated: excessive noise destroys organization, insufficient noise produces rigidity, and the intermediate regime living systems occupy is a defining feature of the organizational phase.
 
These properties jointly delimit a bounded region in dynamical state space we treat as an \emph{organizational phase} \citep{chung2022thermodynamics}. This is distinct from thermodynamic phases defined by static structural parameters: determining whether a system occupies this phase requires dynamical information, causal architecture, algorithmic information processing, and regulated internal energy flows.

\section{Plasmas as Calibration Exemplars}
 
Complex plasma systems serve as \emph{calibration exemplars} making the organizational phase visible and measurable, and as \emph{opaque thought experiments} instantiating thermodynamic precariousness and constraint closure in inorganic matter.

\subsection{Why Plasmas?}
As ionized, field-coupled systems, plasmas are generally open, non-equilibrium, and governed by long-range interactions \citep{chen2012introduction, levin2014nonequilibrium}, with multiscale coupling between microscopic motion and macroscopic fields \citep{lapenta2006kinetic}; autopoietic process-patterns within them furnish test environments for life-like complex systems \citep{vladimirov2004dynamic, lozneanu2008cell}. This has precedent: \citet{Hanczyc2010} showed that inorganic oil droplets driven by chemical reactions and Marangoni flows exhibit self-propelled movement and chemotaxis, a ``homeodynamic state from which cognitive processes may emerge.'' Inorganic, physically precarious systems can thus serve as legitimate instruments for probing the boundaries of cognition, provided their dynamics map rigorously onto criteria for minimal agency. Our proposal follows this lineage, using electromagnetic fluxes and charged-particle dynamics where Hanczyc's droplets use chemical gradients and surface flows.
 
\subsection{Macroscopic Dust in Non-Equilibrium}
The empirical foundation draws on \citet{Tsytovich2007}, refined by post-2020 work. In microgravity and ionized-gas environments macroscopic dust grains self-organize spontaneously far from equilibrium, with internal entropy production $d_iS > 0$ from continuous plasma flux.

Because dust-particle interactions are nonreciprocal, the system continuously extracts energy from its environment, a mechanism for proto-metabolic energy sourcing. \citet{Sahu2025} established that varying confinement potential induces clear structural phase transitions, demonstrating plasticity. PK-4 data from the ISS shows microgravity plasmas organizing into field-aligned filaments with states analogous to liquid crystals; these adapt continuously and non-linearly to volatile fluxes, interacting with and restructuring neighboring strings and undergoing topological adaptation via abrupt changes in topological radius.

\subsection{Entropy Production and Thermodynamic Openness}

Entropy production in strongly coupled dusty plasmas at the individual particle level show  obedience to the ECM Steady-State Fluctuation Theorem:
\[
\frac{P(\sigma_\tau = A)}{P(\sigma_\tau = -A)} = e^{A\tau}.
\]
This formally validates the system as a non-equilibrium dissipative structure with quantifiable thermodynamic openness. Because positions and velocities of all particles can be measured simultaneously, dusty plasmas are one of the very few systems allowing direct measurement of entropy production at the kinetic level.
 
\subsection{Sustained Throughput and Broken Equipartition}
Driven plasmas routinely violate equipartition: wave--particle interactions selectively channel energy into particular degrees of freedom, producing persistent anisotropies \citep{Araneda2008, Araneda2015}, so energy is directionally routed through constrained pathways, a physical analogue of metabolic channeling without enzymes or symbolic control.

\subsection{Self-Bounded Structures and Over-Screening}
Plasmas spontaneously generate boundaries (double layers, sheaths, filaments) maintained by the flows they constrain \citep{Fortov2005, Tsytovich2007, MorfillIvlev2009}, and charged particles assemble into long-lived lattices and helical structures persisting only under continued drive. Void formation, dust-free regions bounded by balanced electrostatic and ion-drag forces, constitutes actively maintained topological boundaries mapping onto ``metabolic boundaries'' in recent ALife literature \citep{KliskaNehaniv2024}. These structures realize the system--environment distinction without membranes: the organization defines its own boundary conditions, which in turn shape the organization that generates them, the physical logic of what autopoietic theory calls operational closure \citep{Maturana1980}, which the plasma's tractability makes directly observable.

The mechanism is \emph{over-screening}: in a simple Debye picture each grain's charge is shielded by a monotonically decaying potential, but in dense clouds the continuous flux onto grain surfaces redistributes local ion and electron densities beyond static shielding, so the effective potential overshoots neutrality and develops attractive wells at intermediate distances \citep{Fortov2005, Tsytovich2007}. This replaces repulsive Coulomb interactions with non-reciprocal, distance-dependent attractions, the coupling between electrostatic and flux fields then dictating structural properties. Without continuous flux the grains feel only Coulomb repulsion and disperse, so the structure's existence is functionally tied to its processing of environmental fluxes. This distinguishes dusty plasmas from shallow dissipative structures: where Rayleigh-B\'enard convection is a parametric response to thermal gradients through a fluid of short-lived parcels, dusty plasmas maintain complex topologies through discrete, history-dependent, non-reciprocal interactions between identifiable macro-particles with localized charge histories.
 
\subsection{Constraint Closure and Precariousness}
 
\begin{figure}[htbp]
    \centering
    \resizebox{\columnwidth}{!}{%
    \begin{tikzpicture}[
        box/.style={draw, thick, rectangle, rounded corners, align=center, minimum width=3.8cm, minimum height=1.2cm, font=\small},
        arrow/.style={->, thick, >=Stealth}
    ]
    \node[box] (process) at (0, 0) {Thermodynamic Process \\ \textit{(Continuous Plasma Flux)}};
    \node[box] (work) at (-3, -3) {Thermodynamic Work \\ \textit{(Over-screening Forces)}};
    \node[box] (constraint) at (3, -3) {Formal Constraint \\ \textit{(Helical Dust Structures)}};
    \draw[arrow] (process) to[bend right=20] node[above left, font=\footnotesize] {generates} (work);
    \draw[arrow] (work) to[bend right=20] node[below, font=\footnotesize] {maintains} (constraint);
    \draw[arrow] (constraint) to[bend right=20] node[above right, font=\footnotesize] {channels} (process);
    \end{tikzpicture}
    }
    \caption{Constraint closure in dusty plasmas. The open thermodynamic process (plasma flux) is maintained by formal constraints (helical structures) to produce the work (over-screening) required to stave off entropic dissolution \citep{Kauffman2000, Moreno2015}.}
    \label{fig:constraint_closure}
\end{figure}
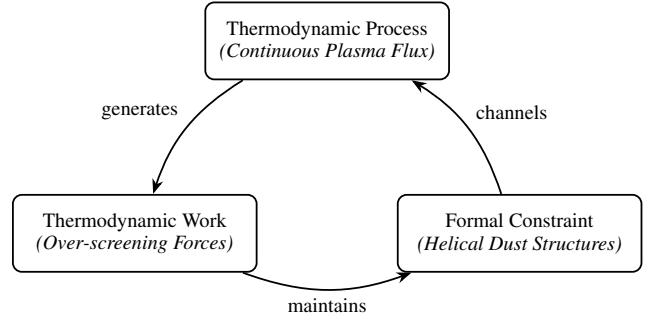
 
Helical dust structures physically instantiate \citeauthor{Moreno2015}'s constraint closure (Figure~\ref{fig:constraint_closure}): the strings act as formal constraints channeling the ionized-plasma flow (the open thermodynamic process), and that channeled flow supplies the over-screening attractions that keep the repelling grains bound in helical shape \citep{Tsytovich2007}. The process generates the constraint and the constraint channels the process, a literal inorganic instantiation of the work-constraint cycle.

A critical caveat: laboratory dusty plasmas rely on externally imposed RF voltage and gas flow for ion streaming \citep{Sahu2025}, so they remain \emph{open} to efficient causation in \citeauthor{Rosen1991}'s strict sense, with self-generated confinement in astrophysical plasmas a testable exception. This locates dusty plasmas precisely between passive dissipation and full organismal closure.

Thermodynamic precariousness most sharply distinguishes the plasma exemplar from computational models. Remove the sustaining flux and helical structures dissolve at once into entropic disorder: no ``pause'' preserves their configuration, no substrate carries their organizational state without continuous drive. A simulation of these physics can be suspended indefinitely \citep{Ahissar2025}; the physical system cannot. This apparently practical difference is philosophically decisive, giving the plasma the existential stakes \citet{Bickhard2000} identifies as the physical root of normativity.
 
\subsection{Structural Memory Without Informational Heredity}
Dusty plasmas exhibit demonstrable physical hysteresis: freezing and melting pressures differ during structural transitions \citep{Sahu2025}, trapped microparticles induce persistent topological defects, and some structures exhibit bifurcation-like events preserving aspects of prior configuration. This topological state memory is grounded in finite dust charging time: the grain retains a physical record of its immediate past. However, dusty plasmas lack a decoupled genotype: no symbolic sequence exerts top-down causal efficacy over phenotypic organization \citep{WalkerDavies2013}. Any structural transfer during bifurcation stays at the level of physical hysteresis and carries no informational heredity, demonstrating minimal individuation \citep{Simondon2020-SIMIIL} without constituting the ``prescriptive information'' required for biological attribution \citep{Benner2010}.

Plasma systems are also inherently noisy \citep{fujisawa2021review}. Under appropriate conditions, stochastic forcing triggers transitions between metastable organized states without degrading them \citep{horsthemke1984noise}, demonstrating that regulated noise sensitivity is a general physical possibility.

Plasmas thus serve as phase benchmarks, realizing the organizational conditions for life- and consciousness-attribution without instantiating cognition or symbolic representation: any engineered consciousness candidate must at minimum satisfy the same physical constraints that plasmas satisfy transparently.

\section{Diagnostic Criteria}
 
We formalize the organizational phase as a bounded region of dynamical state space. These diagnostics function as \emph{admissibility criteria}: they establish whether the physical preconditions for sense-making are satisfied before any higher-level theory is applied.

\subsection{Diagnostics as Phase Constraints}
Let $x(t) \in \mathcal{X}$ denote the system's state trajectory. We define $\mathcal{C} \subset \mathcal{X}$ of admissible organizational regimes such that $x(t) \in \mathcal{C}$ for sustained $t$ is necessary for life- or consciousness-relevant organization. $\mathcal{C}$ is defined conjunctively.
 
\paragraph{Energy throughput and irreversibility.}
The system must sustain non-zero free-energy throughput,
$Q(x) := dF_{\mathrm{in}}/dt \ge Q_{\min}$,
together with positive entropy production,
$\sigma(x) := dS_{\mathrm{env}}/dt > 0$.
These conditions guarantee existential precariousness, excluding equilibrium systems and weakly driven dissipative structures. They are satisfied by dusty plasma helices under continuous drive, where $\sigma_\tau$ obeys the ECM Fluctuation Theorem, and violated by any computational model when paused.

\begin{figure*}[h!]
\centering
\includegraphics[width=\textwidth]{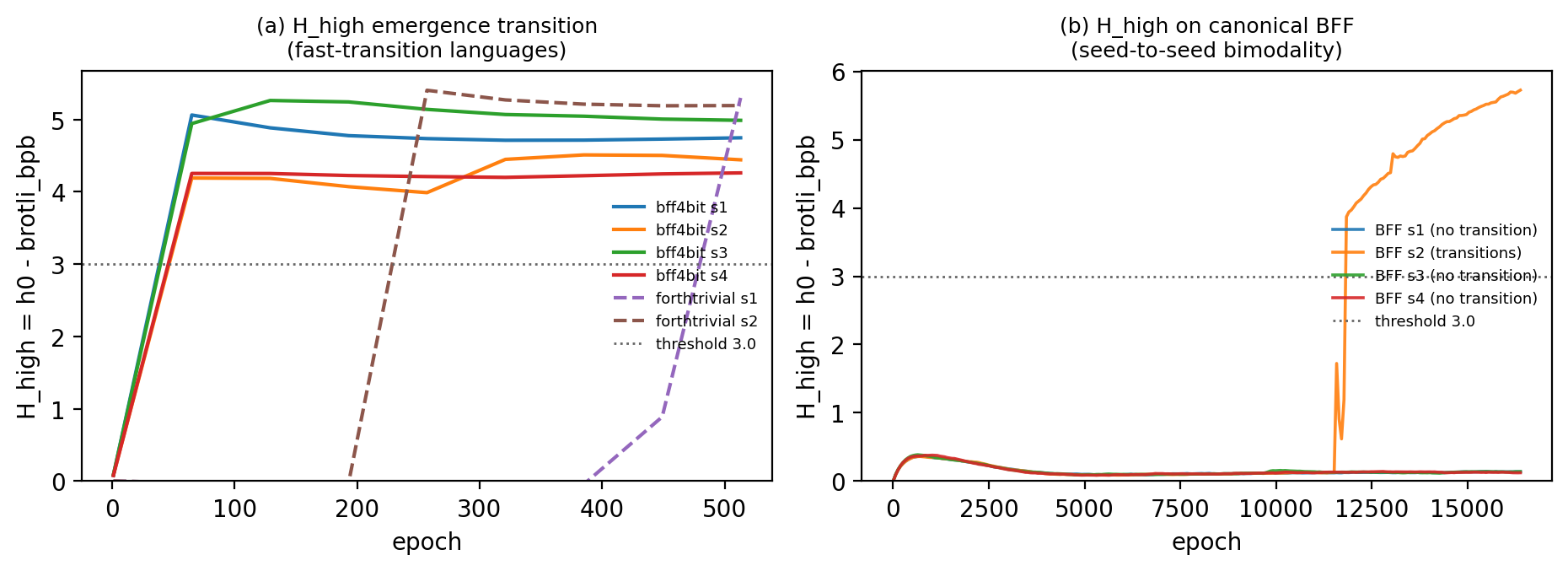}
\caption{\textbf{Informational heredity as an emergence order parameter.} $H_{\mathrm{high}} = H_0 - \mathrm{bpb}$ on the self-replicating-program soup. Transitioning runs rise to a sustained plateau at the onset of self-replication; non-transitioning controls stay near zero. Curves are our own runs on the BFF substrate \citep{alakuijala2024computational}; the transition is bimodal. }
\label{fig:hhigh}
\end{figure*}

\paragraph{Organizational closure.}
Internal dynamics must dominate over external perturbations in maintaining system identity, operationalized as $\Phi_{\mathrm{closure}} := \bigl[\,\mathcal{I}(X_{\mathrm{int}}^{t}; X_{\mathrm{int}}^{t+\Delta t}) / (\mathcal{I}(X_{\mathrm{ext}}^{t}; X_{\mathrm{int}}^{t+\Delta t}) + \epsilon)\,\bigr] \cdot (1 - e^{-\mathcal{I}(X_{\mathrm{int}}^{t}; X_{\mathrm{ext}}^{t})/\epsilon})$, where the numerator measures internal continuity (how predictive the system's own past is of its future) and the denominator external drive. The constant $\epsilon > 0$ serves a dual role: in the denominator it prevents divergence when external influence vanishes, and in the penalty it sets the coupling scale, so when the internal-external mutual information $\mathcal{I}(X_{\mathrm{int}}^{t}; X_{\mathrm{ext}}^{t})$ falls below $\epsilon$ the penalty drives $\Phi_{\mathrm{closure}}$ toward zero, excluding isolated systems that merely conserve an initial state; above $\epsilon$ it saturates and the measure reduces to the ratio of internal continuity to external drive. Sustained closure requires $\Phi_{\mathrm{closure}} \ge \Phi^{\ast}$ with $\Phi^{\ast} \gg 1$, so self-prediction must dominate external drive by a wide margin, capturing the asymmetry central to \citeauthor{Moreno2015}'s framework: the system's own past predicts its future better than the environment's, while it stays thermodynamically coupled to an energy source. The formulation excludes both externally imposed coherence and isolated systems decoupled from throughput.
 
\paragraph{Information maintenance.}
The system must actively maintain information relevant to its persistence. We define $R_{\mathrm{info}} := h_{\mathrm{passive}} - h_{\mathrm{driven}}$, with $h := \lim_{\Delta t \to 0} H(X^{t+\Delta t}\mid X^{t})/\Delta t$, where $h_{\mathrm{passive}}$ and $h_{\mathrm{driven}}$ are conditional entropy rates in passive and driven regimes. A positive $R_{\mathrm{info}}$ indicates that throughput actively reduces information loss, excluding systems that store information without energetic maintenance (crystals, hard drives), which have $h_{\mathrm{passive}} \approx h_{\mathrm{driven}} \approx 0$ and so $R_{\mathrm{info}} \approx 0$ despite retaining information. The measure thus selects for systems whose informational order depends on continued drive, connecting to the plasma's structural state memory. Operationally, $R_{\mathrm{info}}$ can be bounded by comparing phase-space volume under active drive against passive relaxation (RF off): under ion streaming wakefields constrain particles into specific lattice topologies, while passive relaxation expands rapidly into unconstrained Brownian motion.

\paragraph{Regulated noise sensitivity ($\chi_{\mathrm{noise}}$).}
Finally, the system must occupy an intermediate regime of stochastic sensitivity. Let $\eta$ parameterize noise intensity and $\mathcal{O}$ denote an organizational order parameter. We define $\chi_{\mathrm{noise}} := \partial \langle \mathcal{O} \rangle / \partial \eta$, subject to $\partial \sigma/\partial \eta > 0$ (with $\sigma$ the entropy-production rate $dS_{\mathrm{env}}/dt$ defined above),
 $\partial \Phi_{\mathrm{closure}}/\partial \eta \ge 0$, and $0 < \chi_{\mathrm{noise}} < \chi_{\max}$. Here $\partial \Phi_{\mathrm{closure}}/\partial \eta \ge 0$ rules out the destructive regime (added noise may not lower closure), $\chi_{\mathrm{noise}} > 0$ rules out the inert regime (the order parameter must respond to noise at all), and $\chi_{\mathrm{noise}} < \chi_{\max}$ bounds that response. Noise thus aids adaptability without eroding closure.

\paragraph{The admissible organizational phase.}
Collecting these conditions, we define the admissible phase
\[
\mathcal{C} :=
\left\{
\begin{aligned}
x(t) \in \mathcal{X}\ \big|\
& Q \ge Q_{\min},\ \sigma > 0,\ \Phi_{\mathrm{closure}} \ge \Phi^{\ast}, \\
& R_{\mathrm{info}} > 0,\ 0 < \chi_{\mathrm{noise}} < \chi_{\max}
\end{aligned}
\right\}
\]

A system that persistently occupies $\mathcal{C}$ qualifies as a candidate for life- or consciousness-relevant organization. Systems outside $\mathcal{C}$ may be computationally powerful, behaviorally sophisticated, or informationally complex. What they lack is physical grounding.

\paragraph{A separate, measured heredity axis.}
Phase space $\mathcal{C}$ constrains a system's present organization; it is silent on heredity and open-ended evolution, the second pillar of the basic-autonomy framework. We report heredity as a separate, measured axis, and keep it outside the criteria that define $\mathcal{C}$. On a self-replicating-program soup (the BFF substrate of \citealp{alakuijala2024computational}) we track an informational-heredity order parameter $H_{\mathrm{high}} = H_0 - \mathrm{bpb}$: the gap between the order-0 byte entropy $H_0$ and the Brotli compression rate $\mathrm{bpb}$, a computable upper bound on per-byte algorithmic information. A soup carrying repeated, copyable, heritable structure is highly compressible at fixed symbol entropy, so $H_{\mathrm{high}}$ peaks exactly when self-replicators populate the soup. Across replicate runs $H_{\mathrm{high}}$ rises from $\approx 0$ to a sustained plateau at the onset of self-replication, while non-transitioning controls stay near zero (Fig.~\ref{fig:hhigh}); the running code is linked here. We read $H_{\mathrm{high}}$ as a necessary observable signature of an emerging heritable population, not a sufficient definition of heredity: a regular non-replicating pattern can also raise it, and lineage tracing is the complementary check. The axis exposes a double dissociation. By construction the soup carries measured heredity yet, running on equilibrium hardware, cannot host non-equilibrium closure; the dusty plasma carries flux-maintained autonomy yet no informational heredity, showing structural hysteresis and no decoupled genotype. Physical admissibility and biological sufficiency come apart, which is why heredity belongs on its own coordinate.

\begin{table*}[t]
\centering
\footnotesize
\begin{tabular}{@{}lccl@{}}
\toprule
Diagnostic & Convection (driven) & Closure (self-maintaining) & Outcome \\
\midrule
$R_{\mathrm{info}}$ (normalized) & $+0.69 \pm 0.17$ & $-0.69 \pm 0.10$ & separates, opposite sign \\
$\Phi_{\mathrm{closure}}$ (own-past share) & $0.948 \pm 0.004$ & $0.988 \pm 0.013$ & barely separates, confounded \\
$\chi_{\mathrm{noise}}$ (dimensionless gain) & $0.55 \pm 0.04$ & $1.07 \pm 0.20$ & separates, opposite sign \\
\bottomrule
\end{tabular}
\caption{\textbf{Closure-side diagnostics on a controlled physical comparison.} Estimated through a single scalar observable (the spatial standard deviation of the field), the three measures separate a driven Swift--Hohenberg convection pattern from a self-maintaining Gray--Scott structure, but with the sign opposite to expectation. On this observable they track dynamical activity, and the framework therefore makes no claim of a measured closure separation pending an information-theory review. Values are our own runs on the Swift--Hohenberg/Gray--Scott comparison.}
\label{tab:closure-diagnostics}
\end{table*}

The heredity transition above fixes the informational corner on the soup. The closure-side diagnostics we estimate on a controlled comparison: a flux-driven Swift--Hohenberg convection pattern (driven, not closed) against a self-maintaining Gray--Scott reaction--diffusion structure (the closure case), read through one declared scalar observable, the spatial standard deviation of the field, each estimator certified against a closed-form or known-sign answer before any measurement is reported. Through this observable the three measures separate the driven and self-maintaining systems, but with the sign opposite to the pre-registered expectation (Table~\ref{tab:closure-diagnostics}). A scalar observable captures dynamical activity and cannot resolve closure: the self-maintaining structure oscillates, drifts, and divides, and is therefore dynamically more complex and more noise-responsive than a rigid convection roll. The estimator code is certified against closed forms; the open question is the choice of observable. We therefore claim no measured closure separation, and read $\Phi_{\mathrm{closure}}$ as an information-closure quantity, predictive self-determination, distinct from the constitutive self-production of autopoietic closure. The scalar instantiation narrows what we claim empirically while leaving the theoretical criteria intact; a structure-aware observable and an information-theory review of the estimator definitions remain prerequisites before any closure number is reported.

\subsection{Formal Criteria Evaluation}

We evaluate dusty plasmas against these criteria using \citeauthor{RuizMirazo2004}'s dual-pillar framework. The result is a characteristic split: the thermodynamic and organizational prerequisites for basic autonomy are demonstrably satisfied, while the informational and evolutionary criteria remain unsupported. Dusty plasmas exhibit quantifiable energy dissipation via nonreciprocal ion wakes, with entropy production obeying the ECM theorem, and their autopoietic boundary is demonstrated by dust void formation \citep{Fortov2005}, which \citet{KliskaNehaniv2024} show can realize autopoiesis via ``metabolic boundaries'' without a lipid membrane.

The picture changes at efficient causation. Laboratory dusty plasmas rely entirely on external RF voltage and gas flow for the ion streaming that sustains their nonreciprocal dynamics \citep{Sahu2025}; they do not recursively generate their own drive, so closure to efficient causation in \citeauthor{Rosen1991}'s strict sense is unsupported, self-generated confinement in astrophysical plasmas being a testable exception that remains unestablished. The gap widens at the informational level: dusty plasma structures show no sequence-based information replication or channel capacity for open-ended storage \citep{WalkerDavies2013}. A demonstration that could change this would induce a specific sequence of topological defects into a microgravity plasma filament and show a bifurcated offspring copying the sequence with mutual information above baseline, the copied sequence altering the offspring's energy-extraction rate and so establishing top-down informational causation \citep{WalkerDavies2013}. Without heredity, open-ended evolution is foreclosed: dusty plasmas lack the heritable, imperfectly replicated records \citet{Benner2010} identifies as necessary for Darwinian selection.

Because each criterion is physical or information-theoretic, the five are measurable in natural and artificial systems alike and act in engineered systems as design constraints: realizing and stabilizing $\mathcal{C}$ is the primary engineering problem, so the diagnostics are generative as well as classificatory, specifying what must hold of any matter that could bear sense-making.
 
\section{Discussion}

Occupation of $\mathcal{C}$ is necessary but insufficient for life- or consciousness-attribution. The diagnostics delimit when such questions are physically well-posed; they do not answer them. The framework admits falsification: if a digital system on equilibrium hardware scales to human-level cognition without prohibitive energetic costs, or if measures of autonomy (Physical Integrated Information, Causal Emergence) attain brain-comparable values on silicon without free-energy throughput, sustained non-equilibrium organization would prove unnecessary.

A structure-aware coordinate is the next refinement. In a prototype, a localization measure (the inverse participation ratio, which reports how concentrated the field is across cells) did not separate the two systems, because the Gray--Scott self-maintenance here is a distributed, self-replicating lattice with no single localized individual; the count and size of connected components did separate them (Table~\ref{tab:structure}), recovering many small bounded units against fewer larger mode-segments. Granularity correlates with, but does not equal, closure; because the coordinate would alter the central construct, it is future work, with a clean single-individual model such as a Lenia orbium \citep{chan2019lenia} a suitable next test.

\begin{table}[t]
\centering
\footnotesize
\resizebox{\columnwidth}{!}{
\begin{tabular}{@{}lccc@{}}
\toprule
Coordinate & Convection & Closure & Separates? \\
\midrule
Filling fraction (IPR) & 0.65 & 0.68 & No \\
Connected components & $64 \pm 5$ & $192 \pm 16$ & Yes \\
Mean component size (cells) & $52 \pm 4$ & $14 \pm 1$ & Yes \\
\bottomrule
\end{tabular}}

\caption{\textbf{A structure-aware coordinate.} The filling fraction (inverse participation ratio) does not separate the systems, while the count and size of connected components do; granularity correlates with closure.}
\label{tab:structure}
\end{table}

\subsection{Intrinsic Cause-Effect Power}
IIT \citep{Tononi2016} partly aligns with our account, distinguishing what a system \emph{does} from what it \emph{is} and insisting that consciousness depends on intrinsic cause-effect power. Applied to von Neumann architectures, its exclusion postulate (only the maximally irreducible cause-effect structure exists, at a single spatiotemporal grain) forces $\Phi_{\max}$ to localize at individual feed-forward gates, excluding macro-level simulations from unified intrinsic causal power; an exact brain simulation on digital hardware would have $\Phi \approx 0$, converging with our thermodynamic rejection of computational functionalism. The frameworks diverge in grounding: IIT evaluates cause-effect power without requiring thermodynamic precariousness, whereas our diagnostics supply the foundation it presupposes on two levels: intrinsic causal power needs a substrate sustaining it against dissipation, which our conditions specify; and IIT's cause-effect repertoires, defined over counterfactual interventions, require perturbations probed physically, which thermodynamic substrates guarantee through stochastic noise but unprobed digital counterfactuals do not. GWT \citep{Baars1988} faces a complementary limitation: global information availability must be physically sustained against decay.
 
\subsection{Refining Substrate Independence}
Life proves substrate-independent regarding carbon chemistry yet thermodynamics-dependent regarding continuous energy flux. \citet{Thagard2022} shows real-world information processing depends inextricably on energy and so on material substrates, falsifying the organizational invariance underwriting strong computationalism. \citet{MilinkovicAru2025} formalize this as ``biological computationalism,'' where in self-organizing systems the algorithm is the substrate: micro-molecular processes exert direct causal power over macroscopic behavior, a scale-inseparability that digital architectures, with their strict hardware-software abstraction, cannot replicate. \citet{Fagan2025} anchors this by defining encodings as metastable basins of attraction whose separability is enforced by conservation laws, so every irreversible computation incurs a physical cost proportional to the export of a conserved quantity.
 
\subsection{Synthetic, Phased Sense-Making}
The same physical criterion that delimits life constrains intelligence. Under Bickhard's interactivism \citep{Bickhard2000}, the plasma's topological adjustments with varying drive instantiate recursive self-maintenance as dynamical regime shifts executed under persistent threat of dissolution, going beyond passive collisions. AI systems in the equilibrium regime shift computational states indefinitely without those shifts being required for existence, simulating recursive self-maintenance without the normative thermodynamic stakes. Sense-making, on this view, is a physical condition instantiated in matter, prior to any functional implementation. \citet{Lerchner2026} extends this to embodied robotics via the \emph{transduction fallacy}: adding sensors and actuators solves referential symbol grounding \citep{Harnad1990} but does not turn syntactic manipulation into intrinsic sense-making, since the controller still operates on symbols alphabetized by an external mapmaker.
 
\section{Conclusion}
We have shown through a graduated, dual-pillar evaluation across five dynamical diagnostics (sustained free-energy throughput $Q$, irreversible entropy production $\sigma$, organizational closure $\Phi_{\mathrm{closure}}$, active information maintenance $R_{\mathrm{info}}$, and regulated noise sensitivity $\chi_{\mathrm{noise}}$) that dusty plasmas instantiate the thermodynamic prerequisites for basic autonomy while lacking the informational heredity required for open-ended evolution. As opaque thought experiments, these systems physically isolate baseline preconditions from biological sufficiency, confirming that the roots of sense-making are thermodynamic, with physical organization prior to algorithm and syntax; a constraint on the class of systems that can instantiate artificial life and machine minds.

\section{Acknowledgements}
NH gratefully acknowledges Georgii Karelin (Okinawa Institute of Science and Technology, Japan) for his valuable contributions during the initial phases of this work. Sincere thanks are also extended to Prof. Dr. Jaime Araneda Sep\'ulveda (Universidad de Concepci\'on, Chile) for introducing the dynamics of flux-maintained non-equilibrium matter.

\footnotesize
\bibliographystyle{apalike}
\bibliography{references}
 
\end{document}